\documentclass[letterpaper]{article}
\usepackage{aaai}
\usepackage{times}
\usepackage{helvet}
\usepackage{courier}
\begin{document}
%
\title{The Lovelace 2.0 Test of Artificial Creativity and Intelligence}
\author{Mark O. Riedl\\
School of Interactive Computing; Georgia Institute of Technology\\
riedl@cc.gatech.edu
}
\nocopyright
\maketitle
\begin{abstract}
\begin{quote}
Observing that the creation of certain types of artistic artifacts necessitate intelligence,
we present the Lovelace 2.0 Test of creativity as an alternative to the Turing Test as a means of determining whether an agent is intelligent. 
The Lovelace 2.0 Test builds off prior tests of creativity and additionally provides a means of directly comparing the relative intelligence of different agents.
\end{quote}
\end{abstract}

\section{Introduction}

Alan Turing proposed the Imitation Game---later referred to as the Turing Test---as a lens through which to examine the question of whether a machine can be considered to think \cite{turing50}.
The Turing Test was never meant to be conducted; indeed many practical methodological details are left absent
Regardless of Turing's intent, the Turing Test has been adopted as a rigorous test of the intelligence capability of computational systems.
Occasionally, researchers make claims that the the test has been passed. The most recent claim involved a chatbot using simple template matching rules. One of the weaknesses of the Turing Test as a diagnostic tool for intelligence is its reliance on deception \cite{levesque12};
agents that are successful at the Turing Test and the closely related Loebner Prize Competition are those that fool human judges for short amounts of time partially by evading the judges' questions.

A number of alternative tests of intelligence have been proposed including Winograd Schemas \cite{levesque12},
question-answering in the context of a television show, and a robot that gives a talk at the TED conference. 
Bringsjord, Bello, and Ferrucci \shortcite{bringsjord01} 
proposed the Lovelace Test, in which an intelligent system must originate a creative concept or work of art. 
For certain types of creative acts, such as fabricating novel, fictional stories, it can be argued that a creative computational system must possess many of the cognitive capabilities of humans. 

In this paper, we propose an updated Lovelace Test as an alternative to the Turing Test. The original Lovelace Test, described in the next section, is thought to be unbeatable.
The new Lovelace Test proposed in this paper asks an artificial agent to create a wide range of types of creative artifacts (e.g., paintings, poetry, stories, etc.) that meet requirements given by a human evaluator.
The Lovelace 2.0 Test is a test of the creative ability of a computational system, but the creation of certain types of artifacts, such as stories, require a wide repertoire of human-level intelligent capabilities.   

\section{Background}

Ada Lovelace \shortcite{lovelace43} notes that
``the Analytical Engine has no pretensions to {\em originate} anything. It can do {\em whatever we know how to order it} to perform'' (note G).
Turing \shortcite{turing50} refutes the charge that computing machines cannot originate concepts and reframes the question as whether a machine can never ``take us by surprise.''

The original Lovelace Test \cite{bringsjord01} 
attempts to formalize the notion of origination and surprise. 
An artificial agent $a$, designed by $h$, passes the Lovelace Test if and only if:
\begin{itemize}
\item $a$ outputs $o$,
\item $a$'s outputting $o$ is the result of processes $a$ can repeat and not a fluke hardware error, and
\item $h$ (or someone who knows what $h$ knows and has $h$'s resources) cannot explain how $a$ produced $o$.
\end{itemize}

\noindent
One critique of the original Lovelace Test is that it is unbeatable; any entity $h$ with resources to build $a$ in the first place and with sufficient time also has the ability to explain $o$. 
Even learning systems cannot beat the test because one can deduce the data necessary to produce $o$.

{\em Computational creativity} is the art, science, philosophy, and engineering of computational systems that, by taking on particular responsibilities, exhibit behaviors that unbiased observers would deem to be creative. There are no conclusive tests of whether a computational system exhibits creativity. 
Boden \shortcite{boden04} 
proposes that creative systems be able to produce artifacts that are {\em valuable}, {\em novel}, and {\em surprising}. 
Unfortunately, it is not clear how to measure these attributes. 
Boden describes surprise, in particular, as 
the experience of realizing something one believed to be highly improbable has in fact occurred.
Automated story generation is the fabrication of fictional stories by an artificial agent and has been an active area in computational creativity.
The strong story hypothesis \cite{winston11} states that story understanding and story telling play a central role in human intelligence.

\section{The Lovelace 2.0 Test}

We propose a test designed to challenge the premise that a computational system can originate a creative artifact.
We believe that a certain subset of creative acts necessitates human-level intelligence, thus rendering both a test of creativity and also a test of intelligence.

The Lovelace 2.0 Test is as follows: 
artificial agent $a$ is challenged as follows:
\begin{itemize}
\item $a$ must create an artifact $o$ of type $t$;
\item $o$ must conform to a set of constraints $C$ where $c_i \in C$ is any criterion expressible in natural language;
\item a human evaluator $h$, having chosen $t$ and $C$, is satisfied that $o$ is a valid instance of $t$ and meets $C$; and
\item a human referee $r$ determines the combination of $t$ and $C$ to not be unrealistic for an average human. 
\end{itemize}

\noindent
The constraints set $C$ makes the test Google-proof and resistant to Chinese Room arguments. 
An evaluator is allowed to impose as many constraints as he or she deems necessary to ensure that the system produces a novel and surprising artifact. 
For example: ``tell a story in which a boy falls in love with a girl, aliens abduct the boy, and the girl saves the world with the help of a talking cat.'' 
While $C$ does not necessarily need to be expressed in natural language, the set of possible constraints must be equivalent to the set of all concepts that can be expressed by a human mind.
The ability to correctly respond to the given set of constraints $C$ is a strong indicator of intelligence.

The evaluation of the test is simple: a human evaluator is allowed to choose $t$ and $C$ and determine whether the resultant artifact is an example of the given type and whether it satisfactorily meets all the constraints. 
Aesthetic valuations are not considered; the artifact need only be deemed to meet $C$ and 
at need not be better than what an average unskilled human can achieve.

The human referee $r$ is necessary to prevent the situation where the evaluator presents the agent with a combination of $t$ and $C$ that might be extremely difficult to meet even by humans. The referee should be an expert on $t$ who can veto judge inputs based on his or her expert opinion on what is known about $t$ and average human abilities.  

With a little bit of additional methodology, the Lovelace 2.0 Test can be used to quantify the creativity of an artificial agent, allowing for the comparison of different systems. 
Suppose there is a set $H$ of human evaluators, each of which performs a sequence of Lovelace Tests, $k = 1 ... n_i$, such that $|C_k| = k$ and $n_i$ is the first test at which the agent fails to meet the criteria given by evaluator $h_i \in H$. 
That is, each evaluator runs the Lovelace Test where the $k$th test has $k$ constraints and stops administering tests after the first time the agent fails the test. 
The creativity of the artificial agent can be expressed as the mean number of challenges passed: $\sum_{i}(n_i) / |H|$. 
With a sufficiently large $|H|$, one should get a good idea of the capabilities of the system, although future work may be necessary to measure the difficulty of any given set of constraints.



The Lovelace 2.0 Test is a means of evaluating the creativity of an agent with respect to well-defined types of artifacts.
The proposed test can also act as a test of intelligence in the case of types of artifacts that require human-level intelligence.
Consider a limited form of the test: the generation of fictional stories.
Fictional story generation requires a number of human-level cognitive capabilities including commonsense knowledge, planning, theory of mind, affective reasoning, discourse planning, and natural language processing. 
A story generator is also likely to benefit from familiarity with, and able to comprehend, existing literature and cultural artifacts.
%
Currently, no existing story generation system can pass the Lovelace 2.0 Test because 
most story generation systems require {\it a priori} domain descriptions.
{\em Open story generation} partially addresses this by learning domain knowledge in a just-in-time fashion \cite{li:aaai2013}, but cannot yet comprehend nor address complex constraints.


%
The Lovelace 2.0 Test is designed to encourage skepticism in the human evaluators. 
Regardless of whether the human judge is an expert in artificial intelligence or not, the evaluator is given the chance to craft a set of constraints that he or she would expect the agent to be unable to meet.
Thus if the judge is acting with the intent to disprove the intelligence, the judge should experience an element of surprise if the agent passes a challenge. 
The ability to repeat the test with more or harder constraints enables the judge to test the limits of the agent's intelligence.
These features are at the expense of a halting condition---the test provides no threshold at which one can declare an artificial agent to be intelligent. 
However, the test provides a means to quantitatively compare artificial agents.
Creativity is not unique to human intelligence, but it is one of the hallmarks of human intelligence. Many forms of creativity necessitate intelligence. 
In the spirit of the Imitation Game, the Lovelace 2.0 Test asks that artificial agents comprehend instruction and create at the amateur levels.


\bibliography{turingbib}
\bibliographystyle{aaai}

\end{document}